\documentclass[11pt]{article}

\usepackage[margin=1in]{geometry}
\usepackage{graphicx}
\usepackage{subfig}
\usepackage{amsmath, amssymb}
\usepackage{float}
\usepackage{hyperref}

\begin{document}

\title{RoughNet: Mapping Arctic Sea Ice Roughness Using Diffusion-Based Super-Resolution of Satellite Imagery}

\author{
Tessa Cannon$^{1}$,
Michel Tsamados$^{2}$,
Petru Manescu$^{1}$,\\
Thomas Newman$^{2}$,
Christian Haas$^{3}$,
Veit Helm$^{3}$,
Weibin Chen$^{2}$,
Randall Scharien$^{4}$ \\
\\
$^{1}$Department of Computer Science, University College London \\
$^{2}$Department of Earth Sciences, University College London \\
$^{3}$Alfred Wegener Institute Helmholtz Centre for Polar and \\
\hspace{1em}Marine Research, Bremerhaven, Germany \\
$^{4}$Department of Geography, University of Victoria
}

\date{}
\maketitle

\begin{abstract}
Accurate estimation of landfast sea ice roughness is critical for climate modeling and safe Arctic over-ice travel, yet existing approaches rely on costly airborne surveys or sparse in-situ measurements, limiting spatial coverage and operational scalability. Here we show that high-resolution sea ice topography can be reconstructed directly from optical satellite imagery using a conditional diffusion framework. Our approach, RoughNet, learns to map 10 m Sentinel-2 multispectral images to locally normalized 1 m surface elevation residual fields, enabling fine-scale roughness characterization from widely available satellite data. Trained on airborne LiDAR data from two Arctic regions and evaluated on an unseen third Arctic region, the model generalizes across diverse ice conditions and partially reproduces small-scale topographic structure. The best-performing model achieves an out-of-domain root mean squared error of 9 cm while preserving the statistical and spectral properties of the underlying roughness field. These results demonstrate that generative diffusion models can recover physically meaningful surface structure from optical imagery alone, providing a scalable pathway for high-resolution sea ice mapping and roughness estimation in data-sparse environments.

\end{abstract}

Remote sensing is crucial in the Arctic, where sparse populations and extreme environmental conditions limit ground-based measurements. This work specifically leverages generative AI to accurately model sea ice surface roughness (SIR), estimations of which are vital for multiple applications. While definitions vary, we define SIR as the measured elevation deviations from a best-fit plane across a specified region. This study focuses on reconstructing fine-scale surface roughness, capturing elevation variability on the order of centimeters to meters (0.01–5 m). In climate studies, SIR is an essential parameter influencing ice-atmosphere interactions, melt-pond formation, and energy balance within climate models \cite{nolin_arctic_2019}. Beyond climate science, surface features directly impact the indigenous Arctic communities that use sea ice as hunting grounds, as a recreational platform, and as a necessary transportation highway for snowmobile travel. Rougher surfaces translate directly into slower travel, higher fuel consumption, increased equipment wear, and greater risk of accidents. Community engagement research has emphasized the need for readily available information on localized surface conditions which they can use to plan safer, more efficient travel \cite{segal_characterizing_2020}.

Obtaining high-resolution surface measurements to compute sea ice roughness precisely still requires airborne surveys, typically LiDAR collected from crewed aircraft or unmanned aerial vehicles (UAVs). These missions are costly and logistically challenging, making them impractical for frequent, large-scale monitoring. NASA's ATLAS instrument on ICESat-2 provides dense 0.7 m along-track laser sampling, but only along narrow, widely spaced beam tracks with a 91-day repeat window, yielding precise profiles rather than continuous surface maps \cite{farrell_mapping_2020}. Traditional approaches to expand the frequency and coverage of roughness estimates without precise surface measurements have relied on proxy methods to infer roughness from various remote sensing methods, including multi-angle optical methods \cite{segal_characterizing_2020, johnson_mapping_2022} and synthetic aperture radar (SAR) backscatter \cite{macdonald_arctic_2024}. However, these proxy methods are fundamentally limited by the native resolution and sampling geometry of the sensors they rely on. 

Deep learning can significantly improve this process by learning patterns from data to estimate surface roughness. This research aims to leverage satellite imagery to produce more frequent and accurate SIR estimates, capitalizing on the continuous, global availability of remote sensing data from an extensive constellation of satellites. We choose to use generative models because they offer the unique ability to recover fine detail from coarse observations, otherwise known as super-resolution, bypassing the spatial constraints of proxy-based methods. 

We model a mapping 
\[
G: \mathcal{X} \rightarrow \mathcal{Y}_{\text{res}},
\]
where $\mathcal{X}$ represents the multispectral optical satellite observations (Sentinel-2) and $\mathcal{Y}_{res}$ denotes locally demeaned residual topography, enabling prediction of pixel-wise surface roughness fields from satellite imagery. These residual maps can then be used to estimate SIR. 

Deep learning has enabled the prediction of three-dimensional surfaces from optical imagery across computer vision and remote sensing domains. Satellite-based approaches have focused on inferring digital elevation models (DEMs) or refining existing digital surface models (DSMs) using optical or multi-spectral inputs. For example, Stucker et al. trained a convolutional model to refine an imperfect input DSM by regressing a per-pixel height correction using both the DSM and ortho-rectified satellite imagery as conditioning \cite{stucker_resdepth_2022}. Generative deep learning approaches have also increasingly been explored for elevation reconstruction and super-resolution. GAN-based methods have been applied to both inpainting sparse elevation data \cite{y_cai_generative_2023, yan_high_2021} and upsampling coarse elevation data \cite{leong_deepbedmap_2020, demiray_d-srgan_2020}. Ghamisi \& Yokoya introduced IMG2DSM, a conditional GAN framework for generating DSMs from aerial imagery, producing spatially realistic urban structures with an RMSE of approximately 2.5-4 m \cite{ghamisi_img2dsm_2018}. Similarly, Panagiotou et al. used a Pix2Pix-based architecture to generate DEMs from satellite and drone imagery, producing visually detailed terrain, albeit with slightly higher error than baseline models \cite{panagiotou_generating_2020}. While GAN-based methods are capable of generating plausible surface textures, they often suffer from training instability and can struggle to accurately reproduce fine-grained elevation variability.

Diffusion models have emerged as a compelling alternative for generative modeling tasks, offering improved stability and detail preservation compared to adversarial methods. Unlike single-pass generative models, diffusion approaches iteratively refine predictions, making them particularly well-suited for capturing subtle spatial structure. Recent works have begun formulating monocular depth estimation as a generative diffusion problem, significantly reducing error on standard benchmarks \cite{ke_repurposing_2024, duan_diffusiondepth_2023, patni_ecodepth_2024, yang_depth_2024}. However, applications to remote sensing remain limited. Ramirez-Jaime et al. applied diffusion models to super-resolve 3D satellite LiDAR representations, achieving high structural similarity (SSIM $\approx$ 0.9) at 1 m resolution, though relying on emulated data and incurring high computational cost \cite{ramirez-jaime_super-resolved_2025}. Zhou \& Lee proposed a latent diffusion framework to jointly super-resolve satellite imagery and predict depth, achieving RMSE values of approximately 4 meters. This work demonstrated the potential of diffusion priors for preserving both image detail and geometric consistency \cite{zhou_simultaneous_2024}. 

The application of generative deep learning methods in Arctic snow and ice environments presents additional challenges. Machine learning approaches in the Arctic have primarily focused on segmentation \cite{kortum_sar_2024, w_hong_seaicenet_2024} or super-resolution \cite{au_arisgan_2024, tarasiewicz_multitemporal_2023, liu_super_2019, tao_super-resolution_2019}. Arctic landscapes are characterized by low contrast, homogeneous surface appearance, and strong visual similarity to cloud cover, which challenges optical inference across these works. 

Despite rapid progress in deep learning for three-dimensional inference, there remains a clear gap at the intersection of generative modeling and Arctic surface reconstruction. First, no prior work has attempted generative elevation synthesis in Arctic environments, where remote sensing constraints and safety-critical applications make high-resolution mapping particularly valuable. Second, while diffusion models have demonstrated strong performance across image-to-image tasks, they have not been applied to elevation generation from optical inputs in winter Arctic settings. Third, existing pipelines that relate optical imagery to elevation typically decompose the problem into intermediate steps rather than learning a direct mapping from low-resolution imagery to high-resolution surface structure \cite{zhou_simultaneous_2024}. Finally, to our knowledge, no prior approach has produced 1 meter resolution topographic structure directly from 10 meter satellite imagery, making the proposed formulation a substantive advancement beyond current practice.

\begin{figure}[H]
    \centering
    \includegraphics[width=6in]{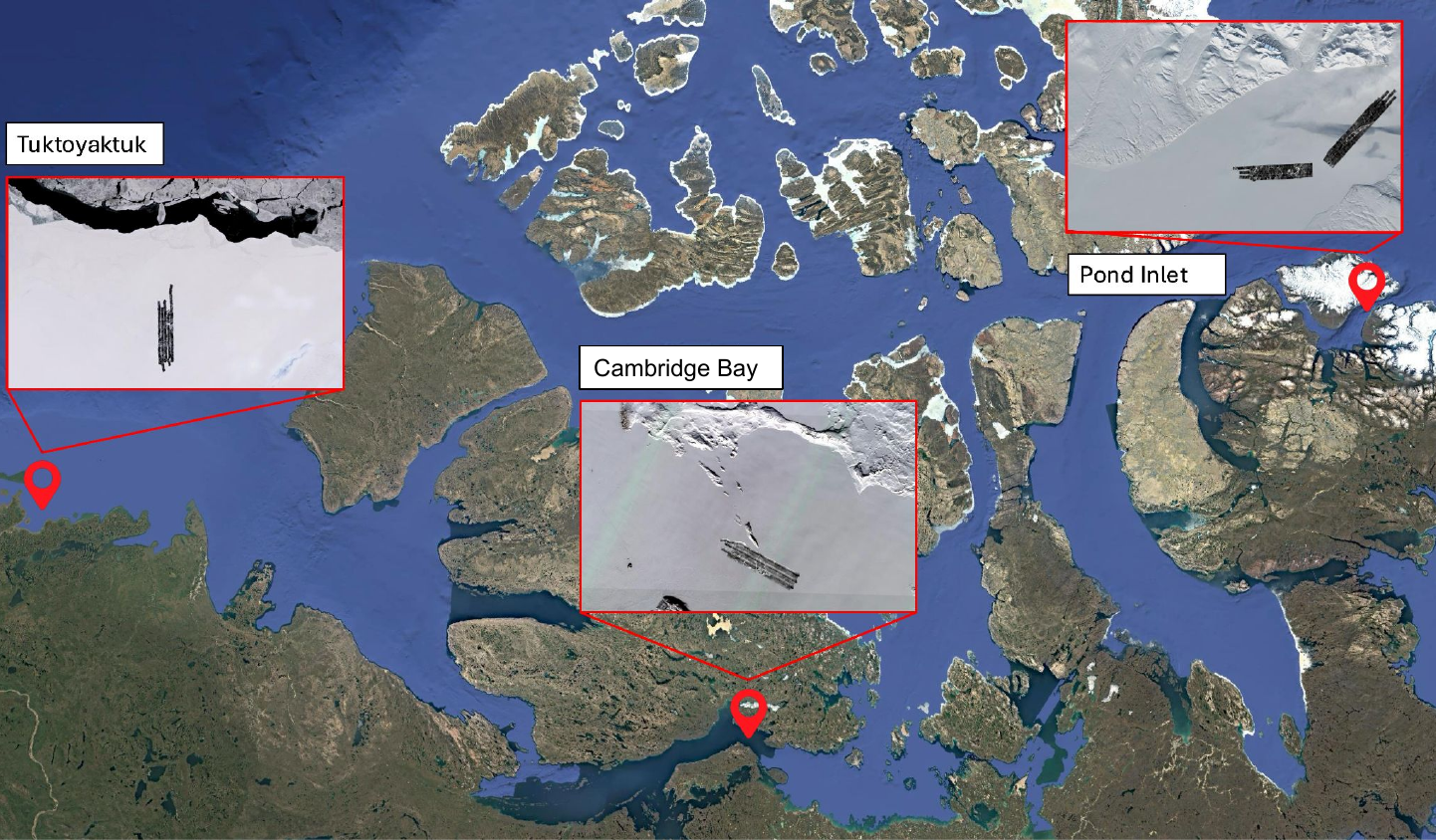}
    \caption{Geographic locations of data collection regions in Northern Canada: Tuktoyaktuk, Cambridge Bay, and Pond Inlet. LiDAR swaths collected from airborne surveys are overlaid on Sentinel-2 satellite views captured in April 2024.}
    \label{fig:lidar_map}
\end{figure}

To address this gap, we propose RoughNet: a conditional diffusion framework that reconstructs meter-scale sea ice topography directly from multispectral Sentinel-2 context. Given stacks of 10 m Sentinel-2 Level-2A images, RoughNet synthesizes 1 m locally normalized surface maps that capture fine-scale texture, enabling high-resolution roughness characterization of Arctic landfast sea ice terrain for navigation and climate applications. We evaluate RoughNet across multiple Arctic regions (Fig. \ref{fig:lidar_map}), examining reconstruction accuracy at both patch and regional scales, and conducting a systematic ablation study of diffusion sampling strategies and noise schedules to identify the optimal model configuration.

\section*{Results}

\begin{figure}[H]
    \centering
    \includegraphics[width=\textwidth]{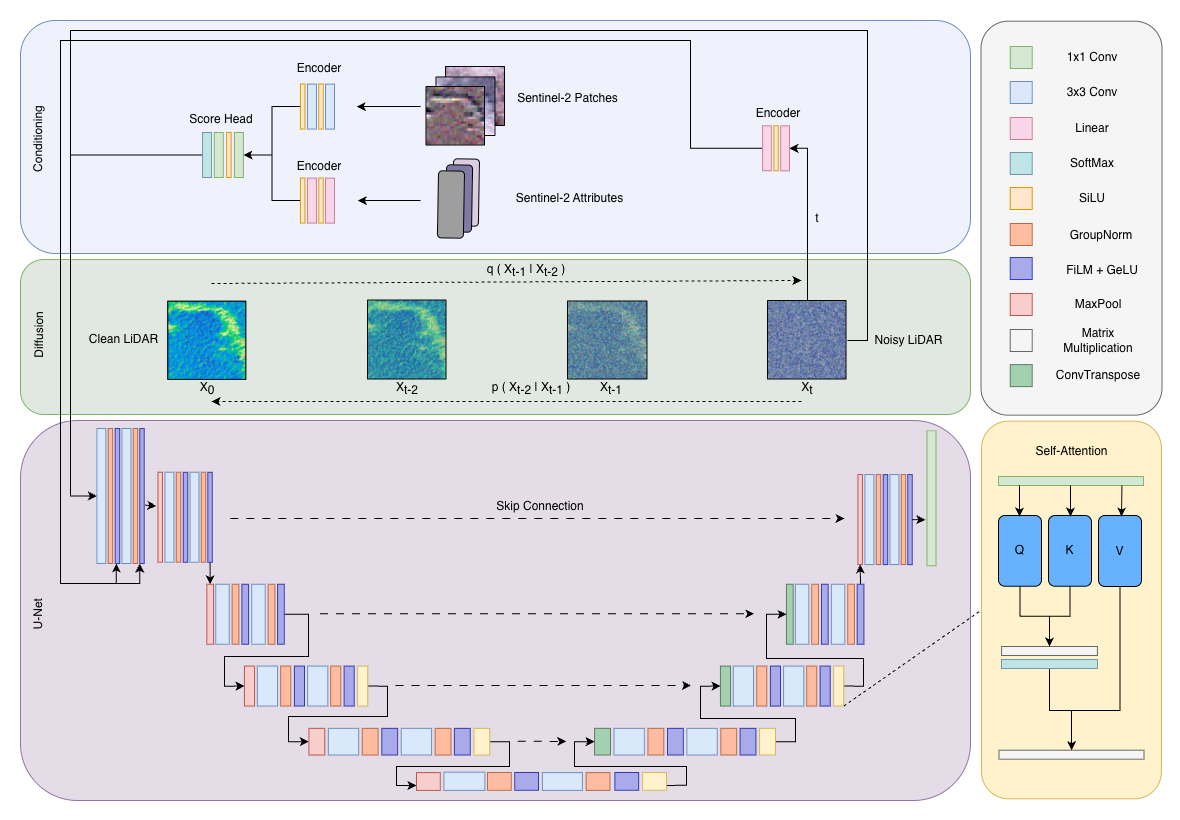}
    \caption{Schematic of the conditional diffusion framework, illustrating the mapping from multi-temporal Sentinel-2 inputs to locally normalized LiDAR residual predictions via a U-Net backbone.}
    \label{fig:model_diagram}
\end{figure}

To learn the mapping from multispectral Sentinel-2 imagery to high-resolution surface texture, we construct a conditional diffusion model based on a U-Net backbone (Fig. \ref{fig:model_diagram}). The model takes six Sentinel-2 images as input and reconstructs locally normalized LiDAR residual patches. The final model was trained for approximately 150 minutes on an NVIDIA GeForce RTX 3090 Ti. The average sampling time for a batch of 8 256$\times$256 pixel LiDAR patches was 30 seconds. We assess reconstruction accuracy and spatial consistency on a hold-out validation set from the training domains, Pond Inlet and Tuktoyaktuk, then test the model’s generalization to new conditions using an unseen geographic test region, Cambridge Bay. 

\begin{table*}[h]
\centering
\caption{Ablation study results of sampling strategy and noise schedule on the validation regions (Pond Inlet + Tuktoyaktuk, equally averaged). Metrics are computed patch-wise in demeaned space. For Bias, the best value is the one closest to zero. Lower is better for all other metrics except ZNCC, where higher is better.}
\label{tab:val_ablation}
\resizebox{\textwidth}{!}{%
\begin{tabular}{lcccccccc}
\hline
Model & RMSE (m) & Bias (m) & $\sigma$ Error (\%) & NAE (°) & ZNCC & JSD & log-PSD RMSE & nRMSE$_\sigma$ \\
\hline
Linear DDPM   & 0.100 & -0.0035 & 20.40 & 1.627 & 0.374 & 0.0923 & 1.686 & 1.050 \\
Linear DDIM   & 0.097 & -0.0036 & 18.16 & 1.626 & 0.396 & 0.0807 & 1.686 & 1.008 \\
Linear PLMS   & \textbf{0.086} & -0.0038 & 26.84 & \textbf{1.428} & 0.458 & 0.0929 & 1.879 & \textbf{0.884} \\
Cosine DDPM   & 0.105 & -0.0017 & 17.95 & 1.831 & 0.395 & 0.0766 & \textbf{0.756} & 1.132 \\
Cosine DDIM   & 0.102 & \textbf{-0.0015} & \textbf{14.35} & 1.783 & 0.404 & \textbf{0.0695} & 0.763 & 1.087 \\
\textbf{Cosine PLMS} & 0.087 & -0.0020 & 17.24 & 1.486 & \textbf{0.467} & 0.0747 & 0.916 & 0.909 \\
\hline
\end{tabular}%
}
\end{table*}

\begin{table*}[h]
\centering
\caption{Ablation study results of sampling strategy and noise schedule on the unseen test region (Cambridge Bay). Metrics are computed patch-wise in demeaned space. For Bias, the best value is the one closest to zero. Lower is better for all other metrics except ZNCC, where higher is better.}
\label{tab:test_ablation}
\resizebox{\textwidth}{!}{%
\begin{tabular}{lcccccccc}
\hline
Model & RMSE (m) & Bias (m) & $\sigma$ Error (\%) & NAE (°) & ZNCC & JSD & log-PSD RMSE & nRMSE$_\sigma$ \\
\hline
Linear DDPM   & 0.102 & -0.0028 & 35.62 & 1.335 & 0.0510 & 0.1175 & 0.876 & 1.495 \\
Linear DDIM   & 0.094 & -0.0027 & 25.48 & 1.308 & 0.0984 & 0.1020 & \textbf{0.866} & 1.358 \\
Linear PLMS   & \textbf{0.083} & -0.0028 & 30.13 & \textbf{1.060} & \textbf{0.1684} & 0.1250 & 0.928 & \textbf{1.142} \\
Cosine DDPM   & 0.129 & 0.0009 & 76.51 & 1.835 & 0.0020 & 0.1296 & 1.329 & 2.010 \\
Cosine DDIM   & 0.109 & \textbf{0.0001} & 38.83 & 1.647 & 0.0527 & \textbf{0.0734} & 1.260 & 1.631 \\
\textbf{Cosine PLMS} & 0.089 & -0.0009 & \textbf{18.36} & 1.170 & 0.1129 & 0.0810 & 1.073 & 1.246 \\
\hline
\end{tabular}%
}
\end{table*}

A model ablation study was performed to determine the optimal diffusion configuration for the proposed task by systematically comparing model performance under different noise schedules and sampling strategies. In particular, we evaluate all combinations of a linear schedule and a cosine schedule against the Denoising Diffusion Probabilistic Model (DDPM), Denoising Diffusion Implicit Model (DDIM), and Pseudo Linear Multi-Step (PLMS) samplers.

No single model is optimal across all metrics and regions (see Tables~\ref{tab:val_ablation},\ref{tab:test_ablation}). While Linear PLMS achieves the lowest RMSE and nRMSE$_\sigma$ across both validation and test regions, Cosine PLMS yields substantially lower $\sigma$ error and Jensen-Shannon Divergence (JSD), indicating improved reconstruction of roughness amplitude and alignment of probability density functions (PDFs). Notably, nRMSE$_\sigma$ values are close to unity across models, indicating that reconstruction errors are on the order of the intrinsic variability of the roughness field. This reflects the fact that RMSE-based metrics penalize point-wise deviations even when the predicted surface preserves correct statistical and structural properties, which are more relevant for characterizing roughness. For the task of roughness prediction, $\sigma$ error and JSD are prioritized over absolute elevation error. Across regions and noise schedules, PLMS tends to outperform DDPM and DDIM across the suite of metrics. These results indicate that both the noise schedule and sampling strategy play a role in preserving fine-scale surface structure, with cosine schedules and higher-order samplers providing the most stable and accurate reconstructions. Cosine PLMS is selected as the final model for subsequent evaluation because of its low $\sigma$ error and JSD, as well as acceptable balance across the remaining error metrics.

The selected Cosine PLMS model achieved an RMSE of $0.09$ m in both validation and the unseen test region (approximately 9 cm), indicating that reconstruction accuracy generalizes well out of domain. Bias was near zero in validation, indicating symmetric errors with no systematic over- or underestimation. $\sigma$ error ($17.24$\% validation, $18.36$\% test) and NAE ($1.49^\circ$ validation, $1.17^\circ$ test) were similarly low across both regions, indicating consistent roughness-amplitude recovery and minimal directional distortion. JSD was likewise low ($0.07$ validation, $0.08$ test), reflecting close agreement between predicted and true elevation distributions. In contrast, ZNCC dropped from a moderate $0.47$ in validation to $0.11$ in the test region, and log-PSD RMSE rose from $0.92$ to $1.07$, showing that fine-scale structure remains imperfectly reconstructed: the model captures overall amplitude and orientation but over-smooths surfaces and misallocates high-frequency spectral power, a systematic rather than region-specific limitation. The absence of degradation in RMSE, $\sigma$ error, and JSD on unseen data indicates robust generalization to novel Arctic terrain, supporting large-scale roughness mapping from satellite imagery.

\begin{figure*}[!t]
    \centering
    \includegraphics[width=6.5in]
    {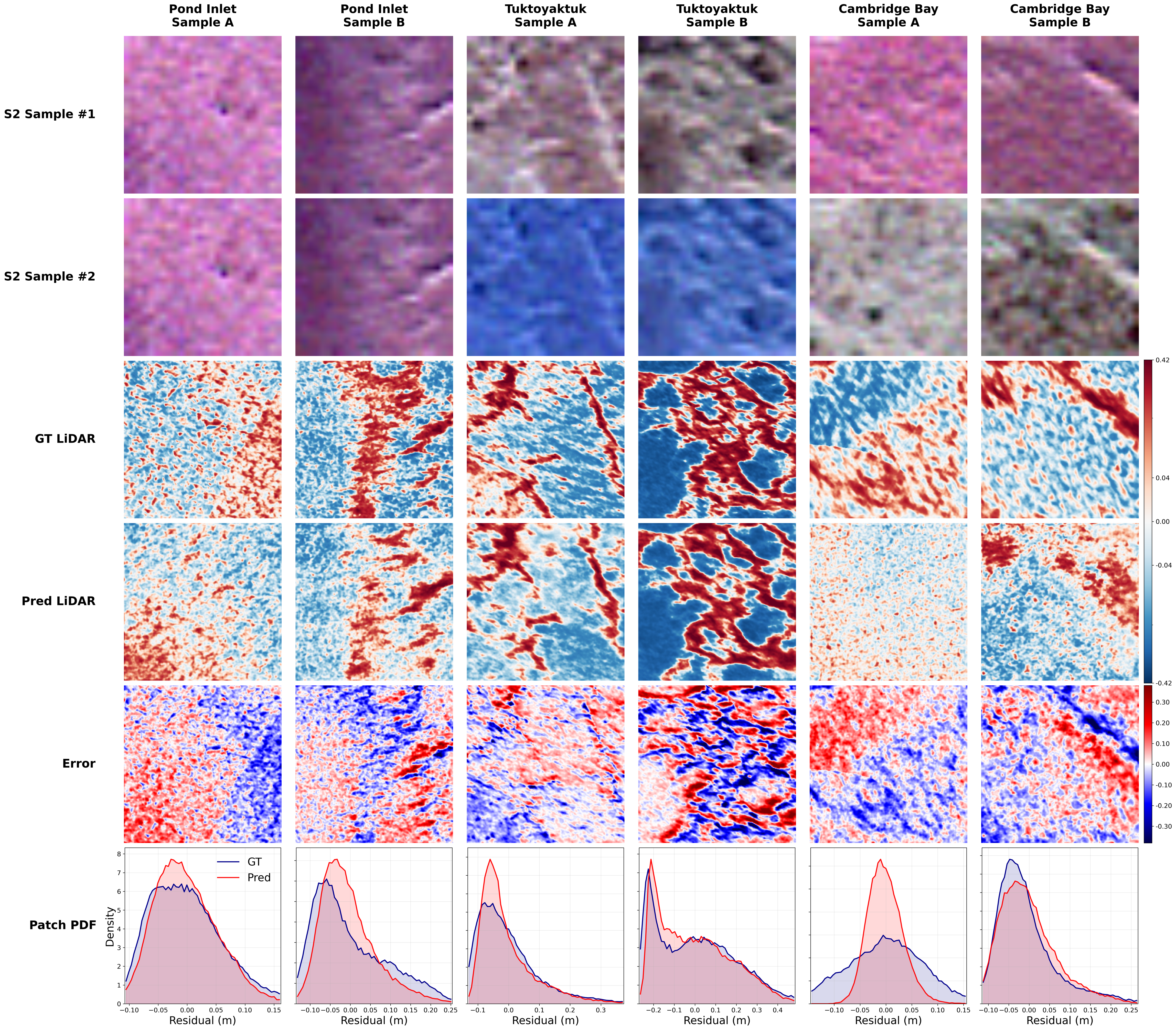}
    % {Figures/patch_reconstructions_pdfs.png}
     \caption{Patch-wise reconstructions from the conditional diffusion model across validation regions (Pond Inlet and Tuktoyaktuk) and the test region (Cambridge Bay). Top rows show two samples of the six Sentinel-2 conditioning inputs; subsequent rows show ground truth (GT) LiDAR, predicted residuals, error (prediction - ground truth), and PDFs of predicted versus ground truth. Patches are visualized in locally demeaned space. }
     \label{fig:patch_reconstructions_pdfs}
\end{figure*}

Visual inspection of patch-wise reconstructions across all three regions reveals strong ability to recover fine-scale topographic detail from optical inputs (Fig. \ref{fig:patch_reconstructions_pdfs}). Visually, predicted patches capture the general structure and resolution of ground truth patches. In Sentinel-2 scenes containing prominent ridges or level ice, the model generally recovers the correct large-scale morphology, demonstrating strong ability to capture coarse structure. However, in areas dominated by low-contrast or repetitive ice textures, predictions tend to produce patterns that do not mirror the specific ground-truth microstructure (see Fig. \ref{fig:patch_reconstructions_pdfs}, Pond Inlet Sample A). Importantly, however, the predicted fine-scale patterns vary across patches, indicating the model is not merely emitting stochastic noise but attempting to condition on the imagery. Additionally, there is no evidence of the model attempting to hallucinate unseen large-scale structures, instead defaulting to fine-scale patterns when uncertain. Error maps show larger residuals near strong features\textemdash largely a consequence of higher true magnitudes in those zones\textemdash whereas finer texture areas have noisy errors randomly distributed around zero. Tuktoyaktuk has the most visually coherent reconstructions likely due to the greater prevalence of ridges and highly variable structure. Out-of-domain Cambridge Bay reconstructions exhibit the highest error, exhibited by significant missing structure in Cambridge Bay Sample A (Fig. \ref{fig:patch_reconstructions_pdfs}).

\begin{figure*}[!t]
    \centering
    \includegraphics[width=\textwidth]{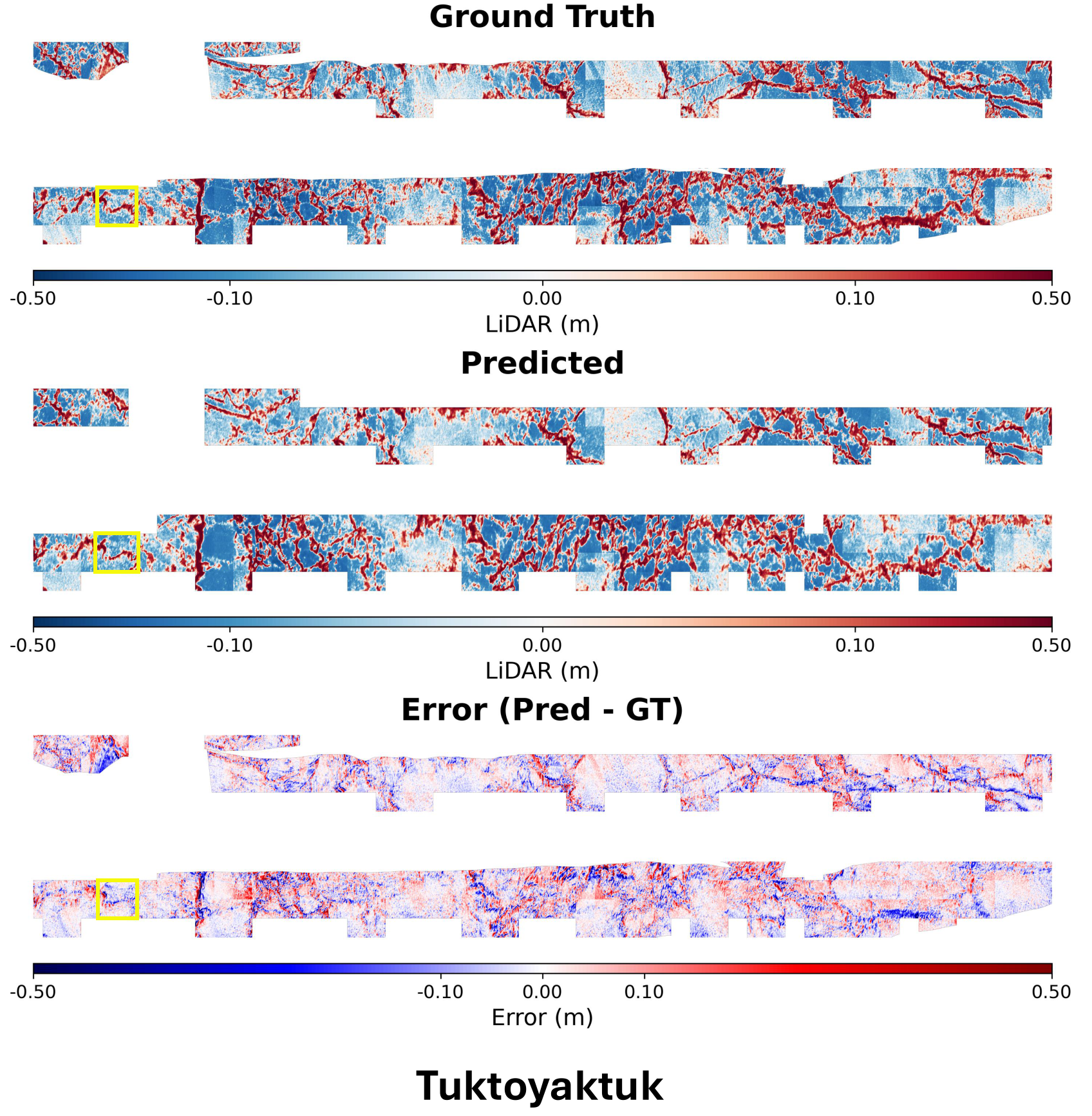}
    \caption{Regional reconstruction for Tuktoyaktuk. Patches are represented in locally demeaned space and mosaicked into a regional representation. This figure illustrates the continuous regional reconstruction specifically for the Tuktoyaktuk region; the corresponding regional mosaics for Pond Inlet and Cambridge Bay are provided in Supplementary Figs. S1 and S2. The yellow bounding box denotes the specific individual patch selected for detailed morphological and statistical analysis in Fig. \ref{fig:patch}.}
    \label{fig:mosaics}
\end{figure*}

To qualitatively assess model performance at the regional scale, validation and test patch sets were processed independently and subsequently mosaicked into continuous maps (Tuktoyaktuk regional reconstruction shown in Fig. \ref{fig:mosaics}, see Supplementary Figs. S1 and S2 for Pond Inlet and Cambridge Bay regional reconstructions). Across all three regions, the reconstructed surfaces exhibit coherent topographic structure that aligns well with the LiDAR references, indicating that the model preserves large-scale spatial organization beyond individual patches. Consistent with the patch-level analysis, Tuktoyaktuk demonstrates the strongest visual reconstruction quality at the regional scale, with particularly effective recovery of prominent features. In contrast, certain elongated, thin ridge structures present in the Cambridge Bay test region are not fully captured by the model. This limitation is likely attributable to the model’s locally constrained receptive field: features that span multiple patches may not be consistently reconstructed without global spatial context. 

As illustrated in Supplementary Fig. S1, the ground truth LiDAR data appears to have fine-scale data artifacts; across LiDAR swaths, there are areas of high-point-density banding, likely the result of mild aircraft roll caused by flying perpendicular to cross-winds, which leads to swath shifts and density gradients in point collection. Because the model never ingests LiDAR at inference, the predicted surfaces are largely unaffected by this collection artifact. This has two implications: (1) regional error metrics are likely conservatively biased where ground truth is degraded, so true performance may be slightly better than reported; and (2) this framework could be used as a tool to flag or mitigate inconsistencies in airborne LiDAR campaigns, since the image-conditioned predictions remain stable where the reference data are locally distorted.

\section*{Discussion}

The superior performance of the cosine noise schedule and PLMS sampler observed in the ablation study is consistent with prior findings on diffusion process efficiency and numerical stability. Linear noise schedules tend to corrupt signal information too rapidly in early timesteps, effectively discarding high-frequency structure before it can be learned \cite{nichol_improved_2021}. In contrast, the cosine schedule preserves a greater proportion of the original signal during early diffusion steps, allowing the model to better capture fine-scale spatial features.

\begin{figure*}[!t]
    \centering
    \includegraphics[width=\textwidth]{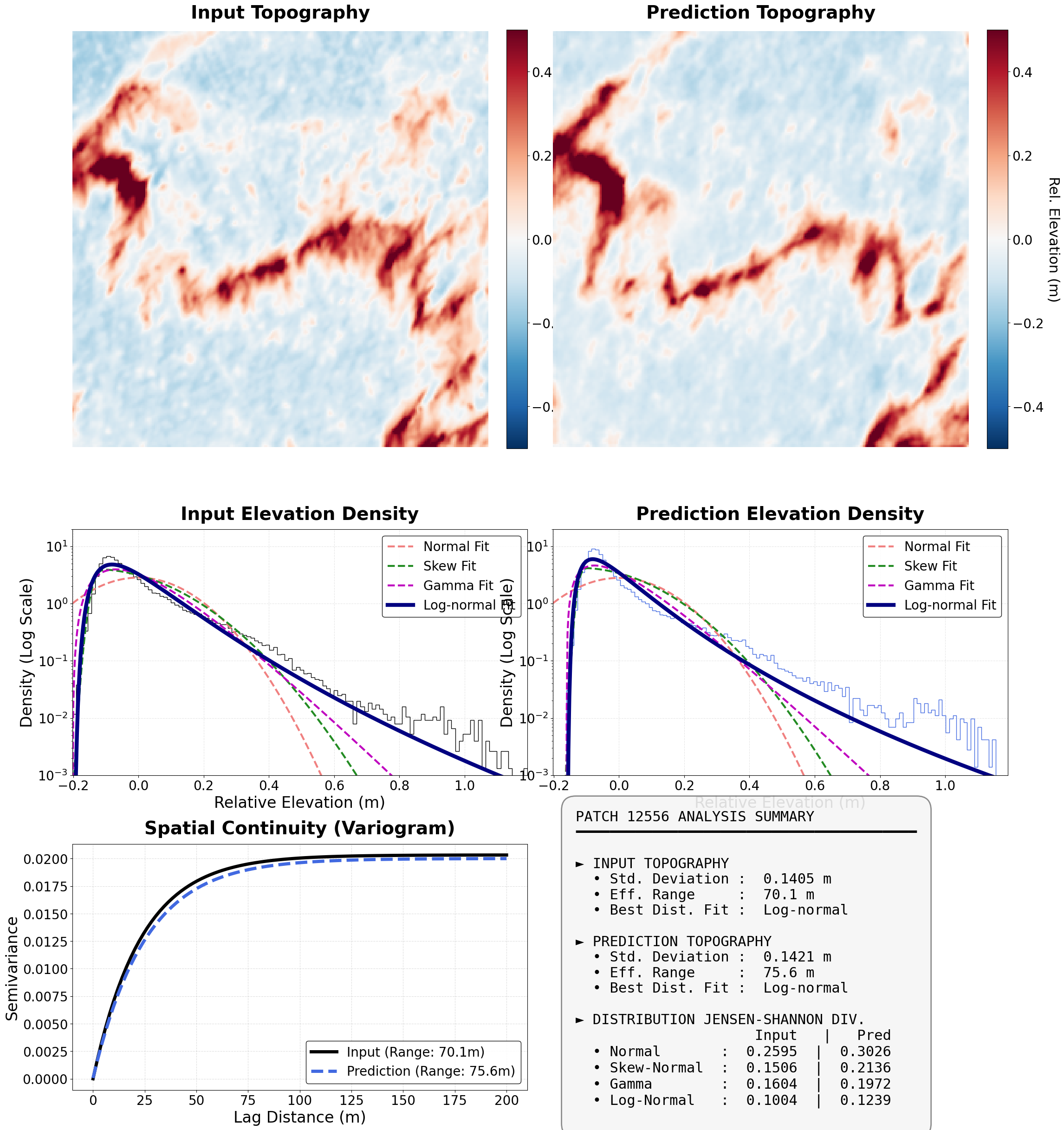}
    \caption{Morphological and statistical patch analysis for a single Tuktoyaktuk patch highlighted in yellow in Fig. \ref{fig:mosaics}. Top panels display the demeaned relative elevation surfaces for the input topography (left) and model prediction (right). Middle panels show the log-scaled elevation frequency distributions overlaid with maximum likelihood estimation (MLE) parametric fits (Normal, Skew-Normal, Gamma, and Log-Normal). Bottom panels provide the spatial continuity comparison via variograms with fitted exponential models (left) and a comprehensive metrics summary block evaluating standard deviation, effective range, and distribution-wise Jensen-Shannon Divergence (JSD) values (right).}
    \label{fig:patch}
\end{figure*}

Similarly, the strong performance of PLMS can be attributed to its higher-order approximation of the reverse diffusion process. Unlike first-order samplers that rely solely on the current estimate during the reconstruction process, PLMS leverages information from multiple previous timesteps to construct a more accurate estimate of the reverse trajectory. This multi-step integration reduces accumulated numerical error and stabilizes sampling, leading to sharper and more physically consistent outputs. Together, these results suggest that both preserving informative signal during forward diffusion and improving numerical accuracy during reverse sampling can enhance high-fidelity cross-modal generation.

% Benchmarking

The most comparable prior work is Zhou \& Lee, which applies diffusion models for joint super-resolution and depth estimation from satellite imagery and reports a depth RMSE of 3.95--4.12 m and an absolute relative error (AbsRel) of 8.5\% \cite{zhou_simultaneous_2024}. Additional close comparisons are Panagiotou et al. \cite{panagiotou_generating_2020}, who reported absolute errors of approximately 270 m (9\% of the elevation range), and Ghamisi \& Yokoya \cite{ghamisi_img2dsm_2018}, who achieved RMSE values of 2.5--4 m in urban DSM reconstruction tasks.

Direct comparison remains imperfect due to fundamental differences in task formulation. While these studies demonstrate the feasibility of cross-modal elevation synthesis, they estimate absolute elevation over large vertical ranges, whereas our model predicts locally demeaned residual topography representing fine-scale surface roughness. Our model achieves RMSE values of 9 cm across all regions at a spatial resolution of 1 m (Tables~\ref{tab:val_ablation},\ref{tab:test_ablation}), corresponding to sub-decimeter accuracy in reconstructing meter-scale surface texture. This represents a substantial improvement in spatial granularity compared to prior work, which typically operates at meter- to decameter-scale vertical errors and coarser spatial resolutions. 

However, normalized error metrics indicate that reconstruction remains high, close to the scale of the signal itself. Across models, nRMSE$_\sigma$ values are generally close to unity (Tables~\ref{tab:val_ablation},\ref{tab:test_ablation}), indicating that the magnitude of reconstruction error is comparable to the intrinsic variability of the roughness field. This reflects the stochastic nature of the task, where multiple plausible realizations of surface topography exist and exact pixel-wise agreement is not expected.

Importantly, pixel-wise metrics such as RMSE and nRMSE$_\sigma$ do not fully capture the objective of this task, which is to reconstruct the statistical and structural properties of sea ice roughness. We therefore emphasize distributional fidelity through the Jensen--Shannon divergence (JSD), which directly measures similarity between predicted and ground-truth elevation probability density functions (PDFs). For roughness characterization, this is a critical property, as many geophysical applications depend on the correct distribution of surface heights rather than exact point-wise correspondence. As shown in Tables~\ref{tab:val_ablation} and \ref{tab:test_ablation}, JSD values for the best-performing Cosine PLMS model are low across both validation (0.07) and test (0.08) regions, indicating strong agreement between predicted and true elevation distributions across spatial scales. 

Under this broader evaluation framework, our results demonstrate that the proposed approach achieves centimeter-scale absolute accuracy while preserving realistic roughness statistics and structure at meter-scale resolution. To the best of our knowledge, this is the first method capable of generating meter-scale sea ice topography from optical satellite imagery, representing a substantial advancement in both spatial resolution and physically meaningful surface characterization.

The morphological and statistical analysis (Fig. \ref{fig:patch}) demonstrates capability for realistic PDF reconstruction. The model faithfully reproduces the log-normal distribution characteristic of natural sea ice and snow elevation, effectively capturing both the heavy tails associated with large ridged features and the tight modal peaks. This structural fidelity is highly consistent with established airborne and ground measurements \cite{landy_facet-based_2019}. Furthermore, the model accurately reproduces the spatial continuity of the surface topography, yielding high-fidelity variograms that reflect empirical correlation lengths known to be governed by underlying ice thickness and wind-driven redistribution \cite{huang_sub-kilometer_2025}. The ability to accurately reconstruct these distributions opens up dual applications: it provides a mechanism to generate synthetic surface topography for altimetry simulations that require realistic roughness parameterizations, while also establishing a foundation for high-resolution mapping to support safe over-ice travel in Inuit localities.

% Limitations and Future Work

While RoughNet demonstrates significant advances, several limitations point to directions for future work.

Training and testing were confined to three geographic regions, limiting generalization beyond their terrain types and surface conditions; broader coverage of distinct features, roughness scales, and illumination would likely improve out-of-distribution performance. The airborne LiDAR used for supervision also contains artifacts and occasional gaps that introduce label noise, which future campaigns could aim to reduce. Sentinel-2 conditioning was also limited to six scenes per region using only the four 10 m bands; higher-quality or more frequent optical inputs, and incorporation of the coarser 20 m and 60 m bands, may provide additional useful spectral context.

Our evaluation metrics, while designed to capture multi-scale structure, cannot fully separate macro- and micro-texture fidelity. Visual inspection remains a necessary complement to quantitative scores, and the practical adequacy of SIR estimates from synthetic surfaces will remain application-dependent.

RoughNet currently assumes six usable Sentinel-2 images per scene, whereas real-world availability varies with cloud cover, polar night, and revisit gaps. Future work should relax this fixed-length conditioning. Inference also takes approximately 30 seconds per batch of eight 256$\times$256 patches, limiting near-real-time use over large areas, and depends on daylight optical imagery, unlike radar sensors such as Sentinel-1 that operate regardless of illumination, albeit at coarser resolution.

The framework is presently restricted to immobile landfast ice, which enables precise spatiotemporal alignment between optical and LiDAR data; extending it to mobile pack ice is conceptually feasible but would require co-location and drift-correction algorithms to account for sea ice motion between overpasses. Finally, stochastic elements in training and sampling (e.g., weight initialization) mean predictions can vary between runs. For safety-critical navigation use, future work should quantify this variance via ensemble or stochastic sampling approaches to produce accompanying confidence maps.

\section*{Methods}

\subsection*{Data Collection} 

Data were collected from three Arctic regions home to Inuit communities in Northern Canada: Pond Inlet (Baffin Island), Tuktoyaktuk (Inuvik Region), and Cambridge Bay (Victoria Island) (Fig. \ref{fig:lidar_map}). These regions were selected for their relevance to Inuit populations, enabling us to train models best suited for the terrain used by indigenous communities for travel. 

Two primary data types are used: airborne LiDAR and satellite imagery. LiDAR was collected during the annual IceBird surveying campaigns using a RIEGL LMS VQ-580 laser scanner (1064 nm, 60° scan angle) mounted on a Basler BT67 aircraft in April 2024 (Pond Inlet: 26-04-24, Cambridge Bay: 18-04-24, Tuktoyaktuk: 16-04-24) \cite{haas_icebird_2024}. Surveys flew at 360 m above ground, yielding a $\sim$350 m swath width and 0.5 m mean point spacing. Raw laser returns were combined with post-processed GNSS trajectories and corrected for aircraft altitude and calibration angles to produce a georeferenced point cloud, with crossover-calibrated elevation accuracy better than 0.1$\pm$0.1 m. A 1 m-resolution DEM was derived using an inverse distance weighting algorithm with a 5 m search radius \cite{franke_iceberg_2025}. Elevations are represented in absolute ellipsoidal height (WGS84), with a $\sim$12.75 m range and $\sim$147 km$^2$ of total survey coverage across the three regions.

The LiDAR data were recorded as three-dimensional point clouds at 1 m resolution, where the first two dimensions correspond to latitude and longitude in WGS84 geographic coordinates and the third dimension represents elevation relative to the WGS84 ellipsoid. The coordinates were reprojected into a locally optimized meter-based coordinate system using a custom Transverse Mercator projection. To remove large-scale elevation trends and emphasize local surface roughness, the raw elevations were converted to residuals by fitting a quadratic surface to each region. Surface fitting was performed using Random Sample Consensus (RANSAC) with a Linear Regression base estimator, providing robustness against outliers. Residuals were computed as the difference between measured and predicted elevation based on the fitted surface, centering the data around zero. Points were excluded if they fell within predefined iceberg polygons, within areas where the laser scanner likely recorded inaccurate measurements (e.g., during sharp turns), or if their residual magnitude exceeded $\pm$10 m relative to the fitted surface. Offsets in absolute elevation vary significantly by region due to differences in sea surface height, motivating the use of local demeaning.

The second data source is multispectral Sentinel-2 imagery from Sentinel-2A/2B, which provide high-frequency global coverage with increased revisit rates at high latitudes. We use the four 10 m optical bands (blue B2, green B3, red B4, near-infrared B8) due to their high spatial resolution and ability to capture fine-scale surface texture. We incorporate multiple Sentinel-2 acquisitions per site, leveraging multi-temporal observations under varying illumination and atmospheric conditions to provide complementary cues (e.g., shadows and contrast) that aid inference of sub-pixel surface structure.

Imagery was retrieved via the Copernicus Data Space Ecosystem API, selecting Sentinel-2 Level-2A scenes within $\pm$14 days of each LiDAR campaign, under the assumption that landfast winter ice conditions remain stable over this window. Cloud contamination was screened manually, as ESA cloud masks are unreliable over snow and ice. Six Sentinel-2 products per region passed filtering and were retained along with their tile-level metadata (solar/viewing geometry, cloud coverage, acquisition date) as auxiliary conditioning information.

\subsection*{Data Preprocessing} 

To enable the model to learn a mapping between LiDAR elevation and optical satellite imagery, LiDAR and Sentinel-2 data must be temporally aligned and geolocated. Because a single Sentinel-2 scene (110$\times$110 km) covers an entire LiDAR collection region, both data sources were divided into smaller, spatially co-registered patches to generate sufficient training examples and align their differing native resolutions. LiDAR patches of 256$\times$256 pixels (1 m resolution) were paired with 26$\times$26 pixel Sentinel-2 patches (10 m resolution) covering the same geographic extent, chosen to balance conditioning context against the number of patches available for training. Patches were extracted using a 256$\times$256 sliding window with 128-pixel stride, then reprojected from the LiDAR CRS to the Sentinel-2 CRS to obtain co-registered multispectral patches. Each patch set was accompanied by acquisition metadata, including solar and satellite viewing geometry (zenith/azimuth means), cloud coverage, and acquisition date.

\begin{figure}[!t]
    \centering
    \includegraphics[width=6in]{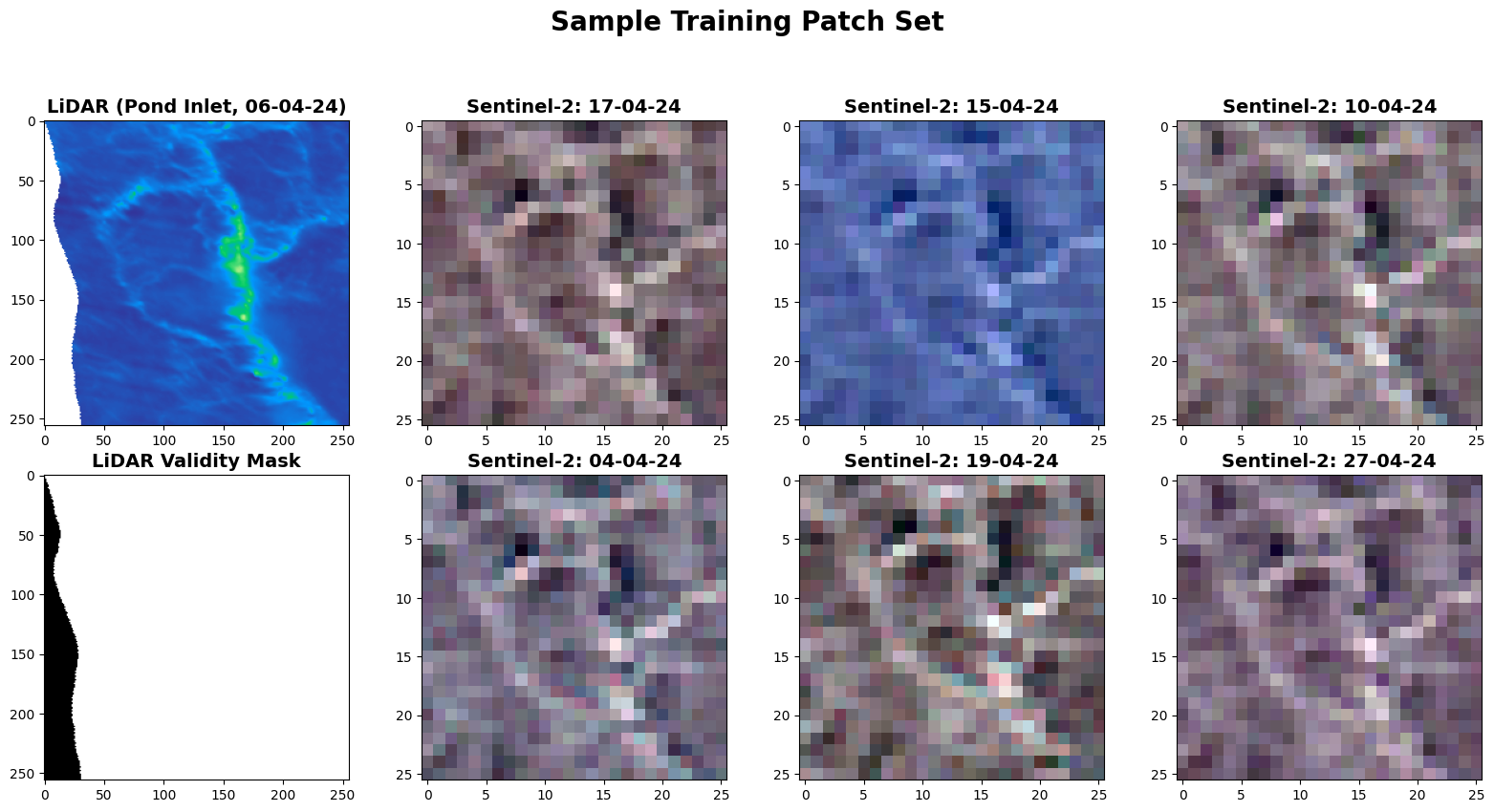}
    \caption{Example patch set from the Pond Inlet region used as training input into the conditional diffusion model framework. Each patch set contains the ground truth LiDAR and its corresponding validity mask, as well as the six Sentinel-2 scenes to be used as conditioning.}
    \label{fig:sample_patch}
\end{figure}

Patch sets with more than 30\% invalid LiDAR pixels were discarded, with a binary validity mask retained for each remaining patch to flag valid elevation values during training. The final training dataset comprises 12,872 patch sets (Pond Inlet: 11,196, Tuktoyaktuk: 1,676), each with one LiDAR patch, its validity mask, six Sentinel-2 patches, and associated metadata (Fig. \ref{fig:sample_patch}). An additional 2,111 patches from Cambridge Bay were reserved for testing.

Sentinel-2 patches were resized to 256$\times$256 pixels via bilinear interpolation and normalized per patch using bandwise shared percentile scaling: values from all temporal views were pooled, clipped to the 2nd-98th percentile range, and rescaled to [0,1] (minimum range safeguard of $10^{-3}$). LiDAR patches were locally demeaned using the mean of valid pixels, removing large-scale elevation offsets so the model learns fine-scale roughness rather than absolute elevation. Training-time augmentation applied random horizontal/vertical flips and 90° rotations jointly across LiDAR, Sentinel-2, and validity masks. Sentinel-2 acquisition metadata (cloud cover, relative acquisition time, viewing/illumination geometry) was encoded as auxiliary conditioning: zenith angles were linearly scaled to [0,1], and azimuth angles sinusoidally encoded to preserve cyclical structure. Patch sets were divided 90-10 into training and validation sets along region-based spatial zones to prevent leakage from overlapping patches.

\subsection*{Theory}

Diffusion models learn to generate data by progressively corrupting samples with Gaussian noise and training a network to reverse this process \cite{sohl-dickstein_deep_2015, ho_denoising_2020}. The forward process is defined as
\begin{equation}
q(x_t \mid x_{t-1}) = \mathcal{N}\!\big(\sqrt{1-\beta_t}\,x_{t-1}, \, \beta_t I \big),
\end{equation}
which admits the closed form
\begin{equation}
q(x_t \mid x_0) = \mathcal{N}\!\big(\sqrt{\bar{\alpha}_t}\,x_0, \, (1-\bar{\alpha}_t) I \big), \qquad \bar{\alpha}_t = \prod_{s=1}^t (1-\beta_s).
\end{equation}

While the reverse posterior $q(x_{t-1} \mid x_t, x_0)$ is Gaussian in closed form, $x_0$ is unavailable at generation time, so a neural network is trained to predict it directly, $\hat{x}_0 = f_\theta(x_t,t)$, following the $x_0$-parameterization for improved stability and interpretability \cite{nichol_improved_2021, he_lotus_2025}:
\begin{equation}
\mathcal{L} = \mathbb{E}_{x_0, t}\Big[ \, || x_0 - \hat{x}_0(x_t, t) ||^2 \, \Big].
\end{equation}
At inference, sampling begins from $x_T \sim \mathcal{N}(0,I)$ and iteratively applies the learned reverse process using DDPM, DDIM \cite{song_denoising_2022}, or PLMS \cite{liu_pseudo_2022}, the latter using multiple prior timesteps to improve trajectory accuracy and sample fidelity.

Conditional generation incorporates auxiliary information $c$ — here, Sentinel-2 image patches and their acquisition attributes — such that $\hat{x}_0 = f_\theta(x_t, t, c)$, enabling reconstruction of high-resolution surface structure from optical satellite imagery.

\subsection*{Model}

A neural network is constructed to perform the reverse diffusion process, taking as input the concatenated noisy sample, conditioning information, and current timestep. We adopt a U-Net backbone \cite{ronneberger_u-net_2015}, whose encoder-decoder structure with skip connections allows the model to combine high-level semantic context with fine-grained spatial detail.

The network accepts the noisy target LiDAR patch concatenated channel-wise with the six Sentinel-2 scenes for local pixel-level conditioning, while timestep information ($t \in [0,999]$) is embedded via multi-layer perceptrons and injected at every block through adaptive group normalization.

Because the number, quality, and atmospheric conditions of Sentinel-2 views vary across regions, the six views are aggregated into a single conditioning map before entering the U-Net. This fusion is permutation-invariant, so the pooled output does not depend on view order: each 4-band patch is encoded by a shared two-layer CNN into a 24-channel feature map, per-view attributes are mapped by a small MLP and broadcast spatially, and a temperature-controlled softmax produces per-pixel weights over the six encoded views, yielding a single 24-channel, attribute-aware map that emphasizes the most informative views.

Within the U-Net, the encoder progressively downsamples features via max-pooling while expanding channels, with skip connections preserving high-resolution cues for the decoder. Each stage uses a double 3$\times$3 convolution block with GroupNorm, FiLM modulation from the timestep embedding, and GELU activations. Scaled dot-product self-attention is applied at the bottleneck and its neighboring layers to capture long-range spatial dependencies. The decoder mirrors the encoder using transposed convolutions and skip concatenations, with a final convolution producing the denoised LiDAR prediction (Fig. \ref{fig:model_diagram}).

During training, locally normalized LiDAR patches are perturbed with Gaussian noise at sampled timesteps, and the network, conditioned on the pooled Sentinel-2 map, predicts the clean sample directly. We optimize a hybrid objective combining per-pixel MSE with L2 gradient regularization on spatial derivatives of the predicted elevation, encouraging structural fidelity, edge preservation, and realistic roughness. The model with the lowest held-out validation loss is retained after each epoch.

\subsection*{Evaluation Methods}

To comprehensively assess model performance, we employ a set of complementary metrics that capture distinct aspects of roughness-field accuracy, spanning local residual reconstruction, statistical realism, and structural fidelity across spatial scales.

Root Mean Square Error (RMSE) quantifies the average deviation between predicted and LiDAR-derived residual fields in meters, serving as the primary measure of pixel-wise reconstruction accuracy. Normalized RMSE (nRMSE$_\sigma$) is computed by dividing the RMSE by the standard deviation of an individual patch, providing a scale-aware measure of error relative to the amplitude of surface height variability. Bias represents the mean signed difference between predicted and true residuals, indicating any systematic asymmetry in the reconstructed surface. RMS height error ($\sigma$ error) compares the standard deviation of the pixel-wise height values, providing a measure of roughness amplitude accuracy. Normal Angle Error (NAE) measures the angular difference between predicted and true surface normals, capturing local slope and orientation accuracy. Zero-mean Normalized Cross-Correlation (ZNCC) evaluates structural similarity between predicted and reference surfaces, reflecting spatial alignment of features independent of absolute magnitude. Average Jensen–Shannon Divergence (JSD) assesses the statistical realism of the reconstructed residual distributions across spatial scales. Finally, log-Power Spectral Density (log-PSD) RMSE quantifies spectral fidelity by comparing the distribution of power across spatial frequencies, capturing discrepancies in fine-scale texture and directional structure.

All metrics are computed at the patch level and subsequently aggregated by region to assess overall reconstruction performance. In addition to these quantitative measures, qualitative inspection of patch-level and region-scale reconstructions is performed to evaluate visual plausibility and structural detail.

\section*{Data availability}
Sentinel-2 data used for this project is publicly available for download using the European Space Agency's Copernicus Data Space Ecosystem. Airborne LiDAR topography data collected during the IceBird campaigns are available upon request from the Alfred Wegener Institute (AWI).

\section*{Code availability}
The code used in this analysis is provided in the GitHub repository available at \url{https://github.com/tessacannon48/RoughNet}. The repository additionally includes figures used in this report as well as sample data patches from the LiDAR datasets to use for process replication. 

\section*{Acknowledgements}

M.T., T.N., and R.S. acknowledge support from the Sikuttiaq project (\url{https://www.cinuk.org/projects/sikuttiaq/}) and the Sikunnguaq project (New Frontiers in Research Fund - International 2023: NFRFI-2023-00588). C.H. acknowledges support from the Sikunnguaq project. 

\section*{Author contributions}

M.T. conceptualised the idea of applying diffusion model interpolation to satellite optical imagery together with LiDAR airborne elevation data, supported the analysis, and provided intellectual support related to polar remote sensing. T.C. developed the RoughNet model architecture, conducted the data preprocessing and machine learning ablation studies, and led the writing of the manuscript. C.H. and V.H. led the IceBird airborne surveying campaigns, collecting and processing the ground-truth LiDAR topography data. P.M. provided technical supervision and expertise on the generative deep learning framework. T.N. assisted with satellite data acquisition and geophysical data preparation. R.S. provided domain expertise on Arctic sea ice surface roughness characterisation. All authors discussed the results and contributed to the editing and review of the manuscript.

\section*{Competing interests}
The authors declare no competing interests.

\bibliographystyle{unsrt}
\bibliography{references}

\clearpage
\newpage

\section*{Supplementary Figures}

\renewcommand{\thefigure}{S\arabic{figure}}
\setcounter{figure}{0}
\begin{figure}[H]
    \centering
    \includegraphics[width=\textwidth]{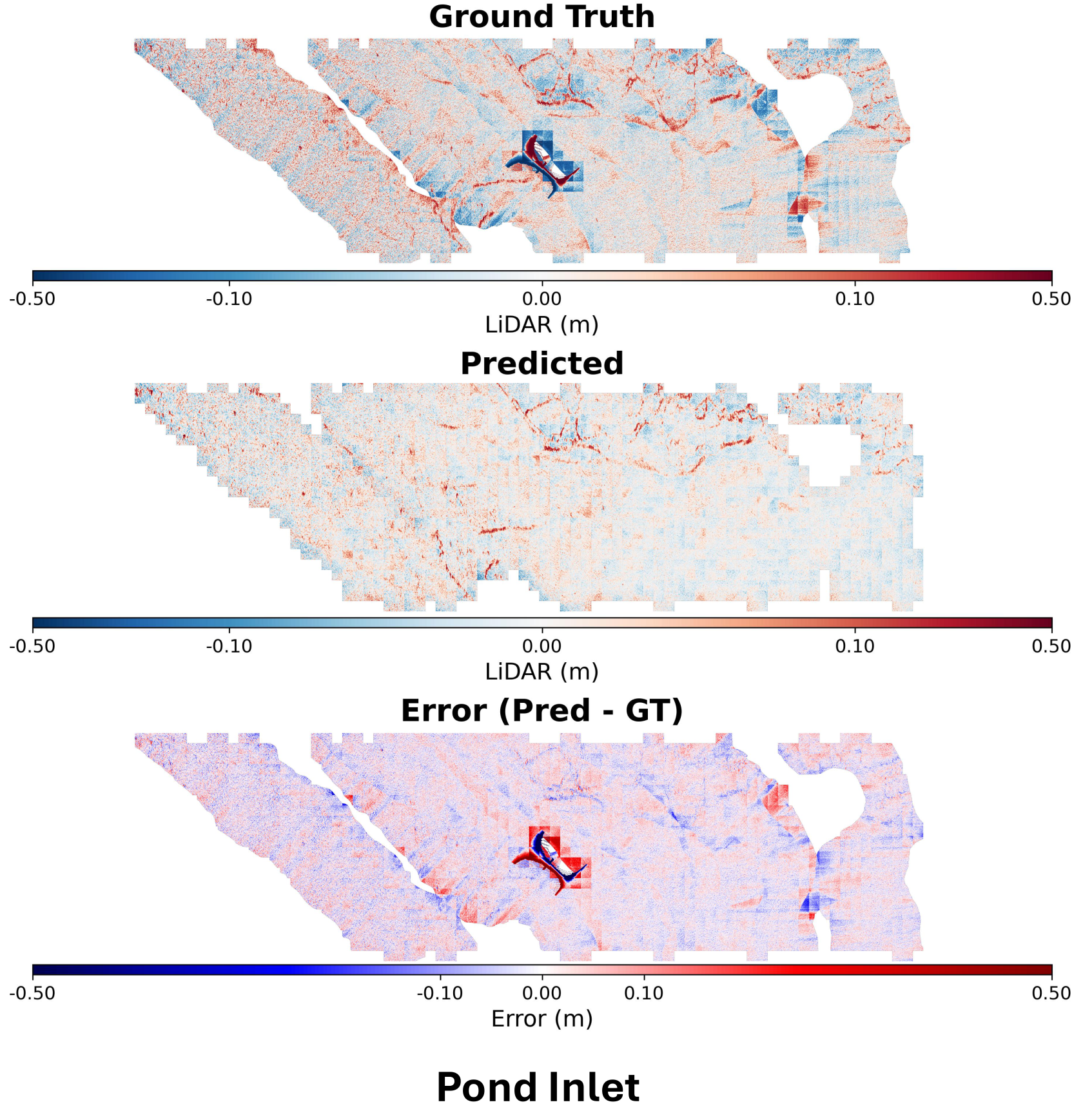}
    \caption{Regional reconstruction for Pond Inlet. Patches are represented in locally demeaned space and mosaicked into a regional representation.}
    \label{fig:supp_mosaics_pond}
\end{figure}

\begin{figure}[H]
    \centering
    \includegraphics[width=\textwidth]{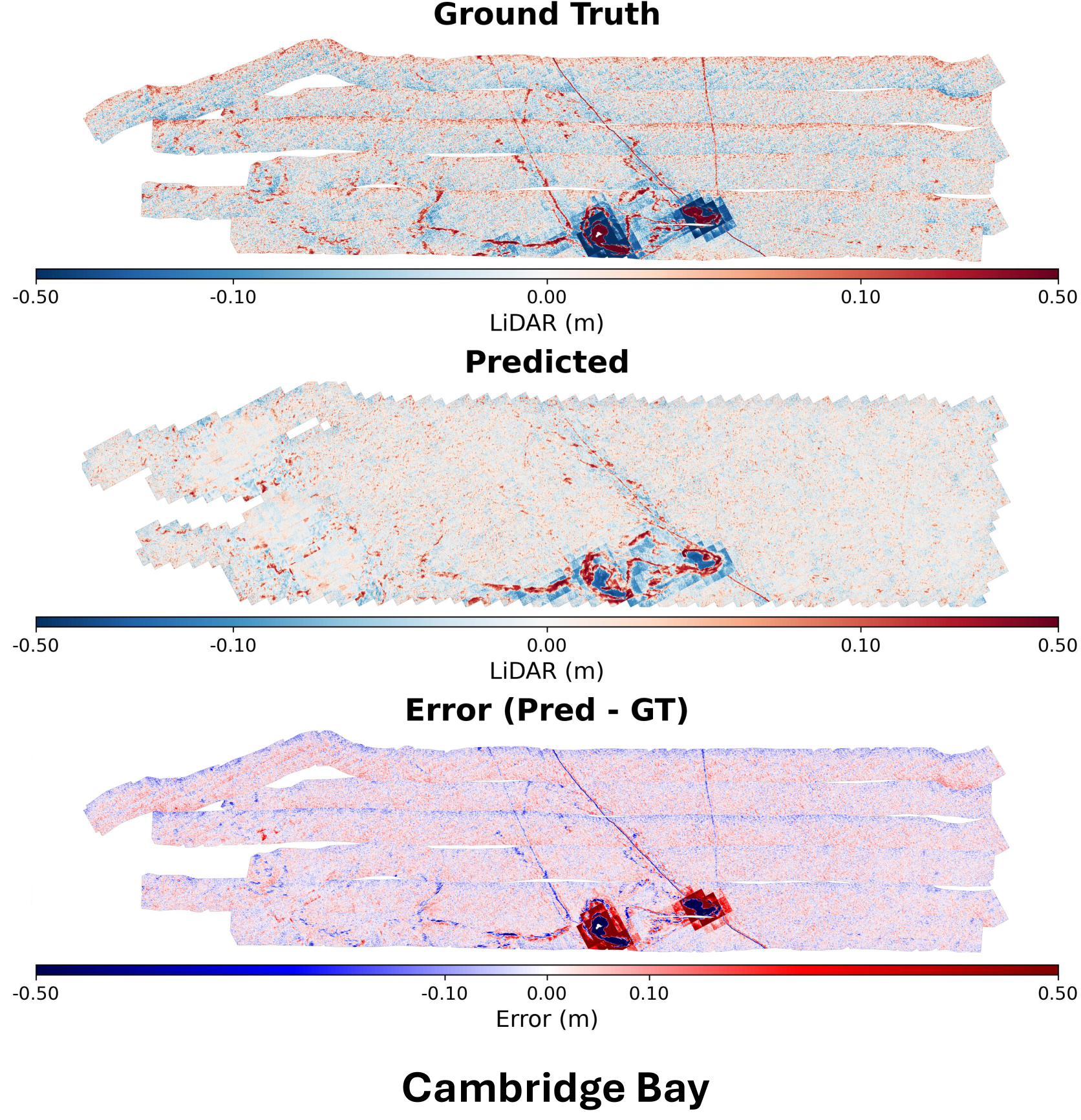}
    \caption{Regional reconstruction for Cambridge Bay. Patches are represented in locally demeaned space and mosaicked into a regional representation.}
    \label{fig:supp_mosaics_cambridge}
\end{figure}

\end{document}